%% file: main.tex
\documentclass[10pt,conference]{IEEEtran}
\input{0-packages}

\input{1-commands}

\input{2-acronyms}
\begin{document}

\input{header}
\maketitle

\begin{abstract}
The limited amount of labeled data for training the Brazilian Sign Language (Libras) to Portuguese Translation models is a challenging problem due to video collection and annotation costs. This paper proposes generating \RM{synthetic videos} \RV{sign language content} by concatenating short clips containing isolated signals for training Sign Language Translation models. We employ the V-LIBRASIL dataset, composed of 4{,}089 sign videos for 1{,}364 signs, interpreted by at least three persons, to create hundreds of thousands of sentences with their respective Libras translation, and then, to feed the model. More specifically, we propose several experiments varying the vocabulary size and sentence structure, generating datasets with approximately 170K, 300K, and 500K videos. Our results achieve meaningful scores of 9.2\% and 26.2\% for BLEU-4 and METEOR, respectively. Our technique enables the creation or extension of existing datasets at a much lower cost than the collection and annotation of thousands of sentences providing clear directions for future works.
\end{abstract}

\IEEEpeerreviewmaketitle

\section{Introduction}

The \gls{libras} is the primary form of communication used by the deaf community in Brazil\cite{leiLibras, decretoLibras}. Despite its official recognition as a means of communication and expression, the linguistic barrier persists, hindering the full inclusion of deaf people in society. In this context, to promote accessibility and inclusion, models that translate \gls{libras} to Portuguese emerge as relevant tools for this purpose.

Two main strategies for translating Sign Languages to spoken Languages are \gls{slr} and \gls{slt}. 
The first consists of extracting meaning from every sign, which implies recognizing each sign individually~\cite{rezende:2021reconhecimento,gameiro:2020signLanguageDataset,passos:2021gait}. 
This strategy can overlook the linguistic properties of sign languages, focusing solely on the visual aspect. 
Another point is that it assumes a direct mapping between sign sequences and spoken language sentences, which is not always valid.
On the other hand, the second strategy aims to generate meaningful sentences in a spoken language given a sequence of signs~\cite{Camgoz:2018:NSLT}. 
Usually, this approach produces results closer to a faithful translation than \gls{slr}-based methods.

The lack of labeled data remains a significant factor in 
the proposal of Brazilian Sign Language translation models ~\cite{dvsilva:2023-libras, zhou:2021improving}.
Although a considerable amount of Libras content is available on the Internet, such as on YouTube channels, many of these videos do not have subtitles or labels indicating what has been signed.
Consequently, to take advantage of these materials, manual translation by a specialist would be necessary, implying an expressive increase in costs and time.
Another possibility is the collection of signed videos for thousands of sentences in a controlled environment, which is yet more laborious and expensive.

Our proposed approach is inspired by~\cite{zhou:2021improving}, which incorporates synthesized massive data for training \gls{slt} models. We create a synthetic dataset by generating sentences from the words available within V-LIBRASIL\cite{rodrigues:2021v}. 
In V-LIBRASIL, for each word, there are, in most cases, three videos of different individuals demonstrating the signs in Libras.
After generating various sentences, corresponding videos of these sentences were created by concatenating the respective short clips of the words presented in each sentence. 
This generated content was used to train an \gls{slt} model~\cite{dvsilva:2023-libras}.

The development of our synthetic dataset and, consequently, the training of an \gls{slt} model represents a significant contribution to the training of Libras translation models without the need for substantial investments in collecting and annotating thousands of videos.
Our primary contributions are summarized as follows:

\begin{itemize}
  \item We propose a new method for creating substantial volumes of data through the concatenation of short video clips containing isolated signals. Additionally, we employ a feature trick to deal with the huge amount of data in an environment with a severe hardware limitation.
  \item We demonstrate the model's ability to learn from \RM{synthetic videos} \RV{concatenated videos} of sentences in Libras with progressive increments in the vocabulary.
  \item We show that results can be improved by increasing the dataset size and variability of the subset of sign short clips. The results provide a clear direction for new research in \gls{slt} for Libras.
\end{itemize}

The manuscript is organized as follows. We present some of the methods used for translating sign language in Section~\ref{sec:relatedworks}. Section \ref{sec:methods} describes the method followed in this work. In Section~\ref{sec:results}, the quantitative and qualitative results are discussed, and finally, in Section~\ref{sec:conclusion}, the conclusions about the experiments and the directions for future work are presented.

\section{Related works}
\label{sec:relatedworks}

Zhou et al.~\cite{zhou:2021improving} proposed an approach to enable the extension of datasets through a mechanism of multiple texts based on gloss videos. The study demonstrated the effectiveness of this synthetic data generation mechanism through experiments. \RV{It differs from our work by using massive spoken language texts to increment its training and dataset, unlike our approach where we created sentences for training the model. Additionally, they used an original approach called Sign Back-Translation.}

Chen et al.~\cite{chen:2023twostream} utilizes two different data streams for model creation: RGB videos and keypoint sequences. They highlight the importance of incorporating domain knowledge in understanding sign language through keypoints.
Several approaches are proposed for the interaction of the two streams, such as bidirectional lateral connection and frame-level self-distortion. 
That work also demonstrates the model's functionality for both \gls{SLR} and \gls{SLT}. The study of Mo Guan et al.~\cite{guan:2024multistream} presents the Multi-Stream Keypoint Attention Network, a novel approach for sign language recognition and translation. 
The model decouples keypoint sequences into four distinct streams: left hand, right hand, face, and full body.
Each stream focuses on specific aspects of the skeletal sequence. 
The approach also employs keypoint fusion strategies and attention mechanisms between the different streams to enhance the interaction and interpretation of gestures in sign language.
This new approach has achieved state-of-the-art performance on translation tasks based on benchmarks.

These works share \RM{the use of} the same datasets, such as RWTH-PHOENIX-Weather 2014~\cite{Camgoz:2018:NSLT}, which enable the training and evaluation of models for other sign languages. 
However, their adoption for Libras is not straightforward due to the absence of datasets containing glosses, sentences, and videos.

Silva et al.~\cite{dvsilva:2023-libras} presented the first \gls{SLT} proposal for Libras using a dataset based on the translation of the Bible.
The work had limitations regarding the results and faced difficulties due to the complexity of the Bible's vocabulary. 
In this paper we demonstrate that synthetic content generated in Libras can be used for training models, facilitating learning.

\section{Methodology}
\label{sec:methods}

In this section, we introduce in detail all the steps followed in this study, comprising data acquisition, pre-processing, experimental configuration, model architecture, \RM{hardware,} and evaluation metrics used. 

\subsection{Data acquisition and pre-processing}
\label{subsec:dataacquisition}

\begin{figure}[!t]
\centering
\includegraphics[width=2.8in]{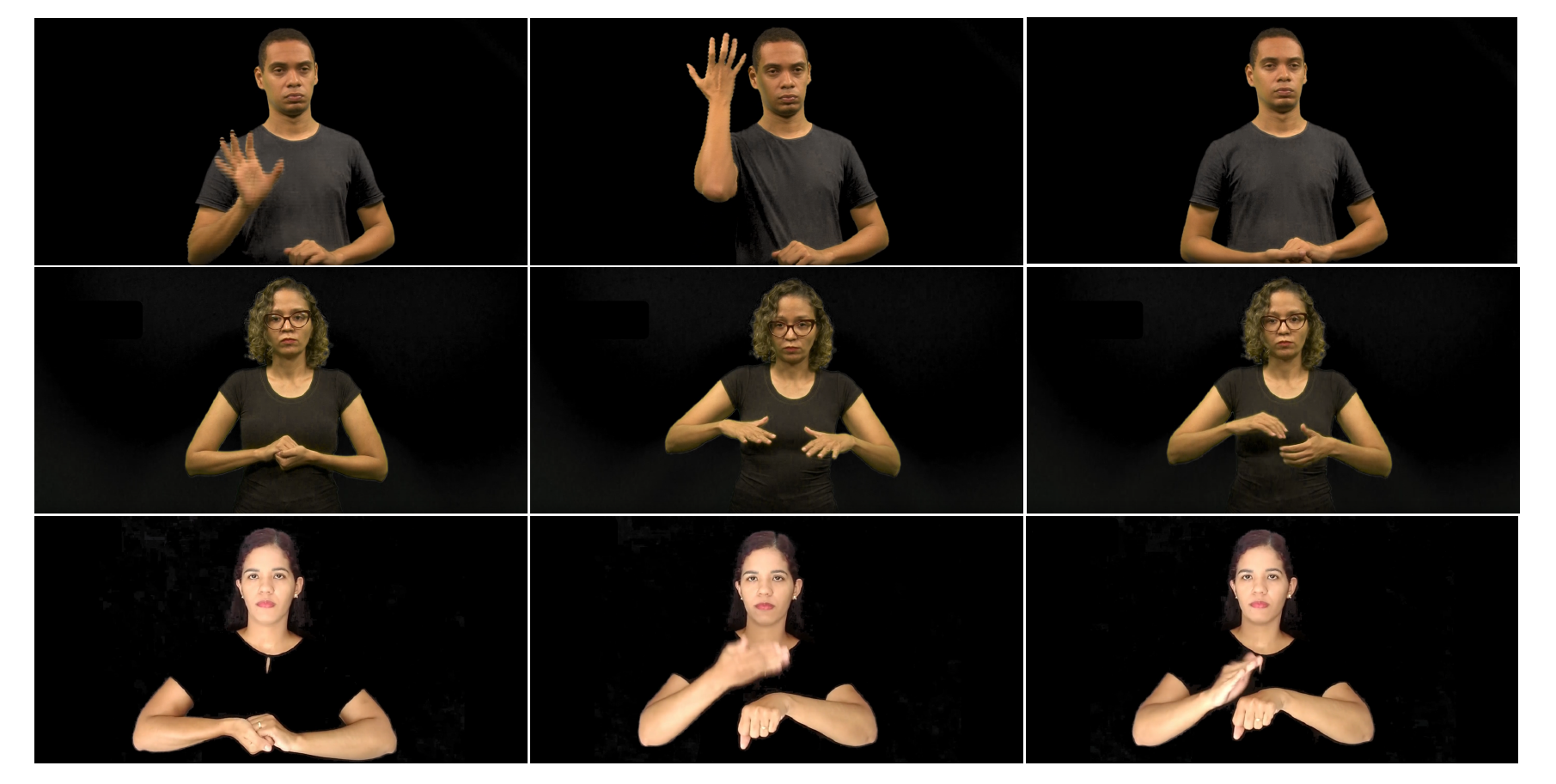}
\caption{Sign language videos from different words of V-LIBRASIL dataset. \RV{The first, second, and third rows present images from the sign videos for the words ``tree", ``depend", and ``train", respectively.} }
\label{fig:v-librasil}
\end{figure}

\paragraph{Dataset} V-LIBRASIL is a \gls*{libras} dataset created in  \cite{rodrigues:2021v} and composed by 1{,}364 signs interpreted at least by three people by sign. 
The dataset consists of 4{,}089 sign videos, recorded at a chroma key environment as illustrated in Fig.~\ref{fig:v-librasil}. 
Each video represents a sign from \gls*{libras} corresponding to a word from Portuguese. The videos are available at their official website\footnote{\url{https://libras.cin.ufpe.br/}}.

\paragraph{Scrapping}
despite the ease of accessing videos, the file names do not indicate which sign is presented or who the interpreter is. Additionally, the relationship between the video file name and the sign is also unclear. To correctly identify which sign corresponds to each video, scraping the sign page and correlating files with signs was necessary. Other difficulties were identified, such as the absence of some videos and the lack of standardization in ordering interpreters by sign, which we checked manually in this study. The main scripts of this study are going to be available at the repository of this work\footnote{\scriptsize{https://github.com/DavidVinicius/concatenating-videos-for-sign-language-translation}}.

\paragraph{Word labeling}
as is described in Section~\ref{subsec:experimentalconf}, we need the grammatical class for each word to construct sentences with a minimum semantic structure. 
Initially, we chose four grammatical classes that appeared most frequently within V-LIBRASIL: nouns, verbs, adjectives, and adverbs. Each word was translated into English, and the grammatical class was determined using the NLTK library~\cite{nltk:2002}.
After this procedure, we found 773 nouns, 225 verbs, 216 adjectives, and 35 adverbs.
The other grammatical classes found were not considered.

\paragraph{Video augmentation}
we applied augmentation techniques from~\cite{vidaug:2018} to provide more variability in the videos during training. 
Six augmentation types were generated for each video: upsample, downsample, horizontal flip, horizontal flip with downsample, and horizontal flip with upsample. 
During training, two types of augmentation were randomly chosen for each sentence from each interpreter.

\paragraph{Feature extraction}
we extract features using the \gls{i3d}~\cite{carreira:2018}, a model widely adopted for action recognition~\cite{xing:2023svformer}, video captioning tasks~\cite{estevam:2021dense}, sign language translation~\cite{Camgoz:2018:NSLT} and many other tasks.
The \gls{i3d} effectively handles temporal and spatial information within video sequences by employing three-dimensional convolutional filters.
This method allows for extracting motion-specific features alongside the static characteristics found in individual frames. 
We create a stack of features from subsets of 10 frames and utilize the RGB and Optical Flow streams of \gls{i3d}.
Each frame was resized so that its shortest side was 256 pixels. 
Next, the center region was cropped to produce $224 \times 224$ pixel frames. 
Finally, the optical flow was estimated using the PWC-net model~\cite{sun:2018pwc}.

\paragraph{Feature trick}
generating the long videos by concatenating the short video clips is straightforward.
However, due to the hardware limitations, we could not treat the weights of \gls{i3d} as learnable parameters. An alternative is to pre-compute the features for the sentence videos. However, considering smaller datasets with sizes from 30,000 to 40,000 sentences, the processing time was around 2 to 3 days with our resources (see \RM{Section~\ref{subsec:hardware}}\RV{\ref{sec:results}}).
This extended processing makes the execution of experiments unfeasible, resulting in a considerable waiting period before the training. 
Considering this limitation, we pre-computed all the feature stacks of each short video clip before the concatenation because several videos shared identical content (e.g., the same sign from the same interpreter and the same augmentation), and the feature stacks are also identical in those cases.
This approach led to a significant improvement. 
What previously took days was reduced to mere hours, and no differences were observed in the experimental results.

\subsection{Experimental configurations}
\label{subsec:experimentalconf}

Our goal is to evaluate whether the model can learn to translate sequences from \gls*{libras} signs into Portuguese. 
Therefore, it is essential to define the rules for the selection of these signs.
In this study, we propose two different configurations: the first, named \gls{sf}, and the second, named \gls{rf}.
In the \gls{sf} configuration, we aim to generate sentences with some semantic meaning.
To address this obstacle, we propose a fixed sentence structure that would be as meaningful as possible and that would utilize the four grammatical classes according to 
\begin{equation} 
\label{sentenceStructure}
 Sentence = Noun \oplus Adjective \oplus Verb \oplus Adverb,
\end{equation}
\noindent where $\oplus$ is the concatenation operator.

On the other hand, in the \gls{rf} experiment, the words do not have a fixed position in the sentence and can appear in any order. 
With this experiment, we aim to determine if the model could learn at the signal level rather than simply memorizing a fixed structure.
\RV{For both experiments, three different tests were conducted with varying numbers of words.}

\begin{figure*}[!t]
	\centering
	\captionsetup[subfigure]{captionskip=-0.25pt,font={scriptsize},justification=centering}
	\vspace{0.35mm}
	\resizebox{0.70\linewidth}{!}{
            \subfloat[][]{
			\includegraphics[width=0.45\linewidth]{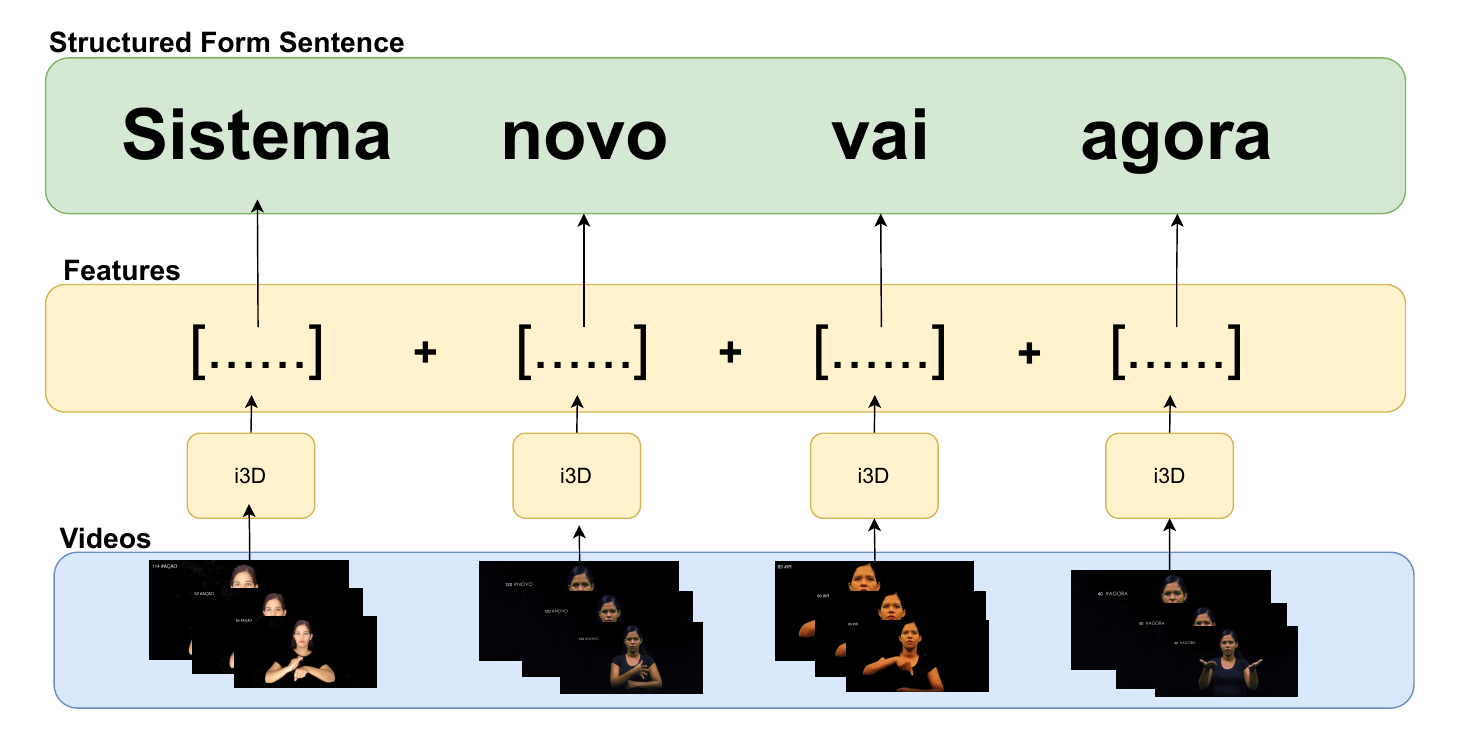} 
		}
            \subfloat[][]{
			\includegraphics[width=0.45\linewidth]{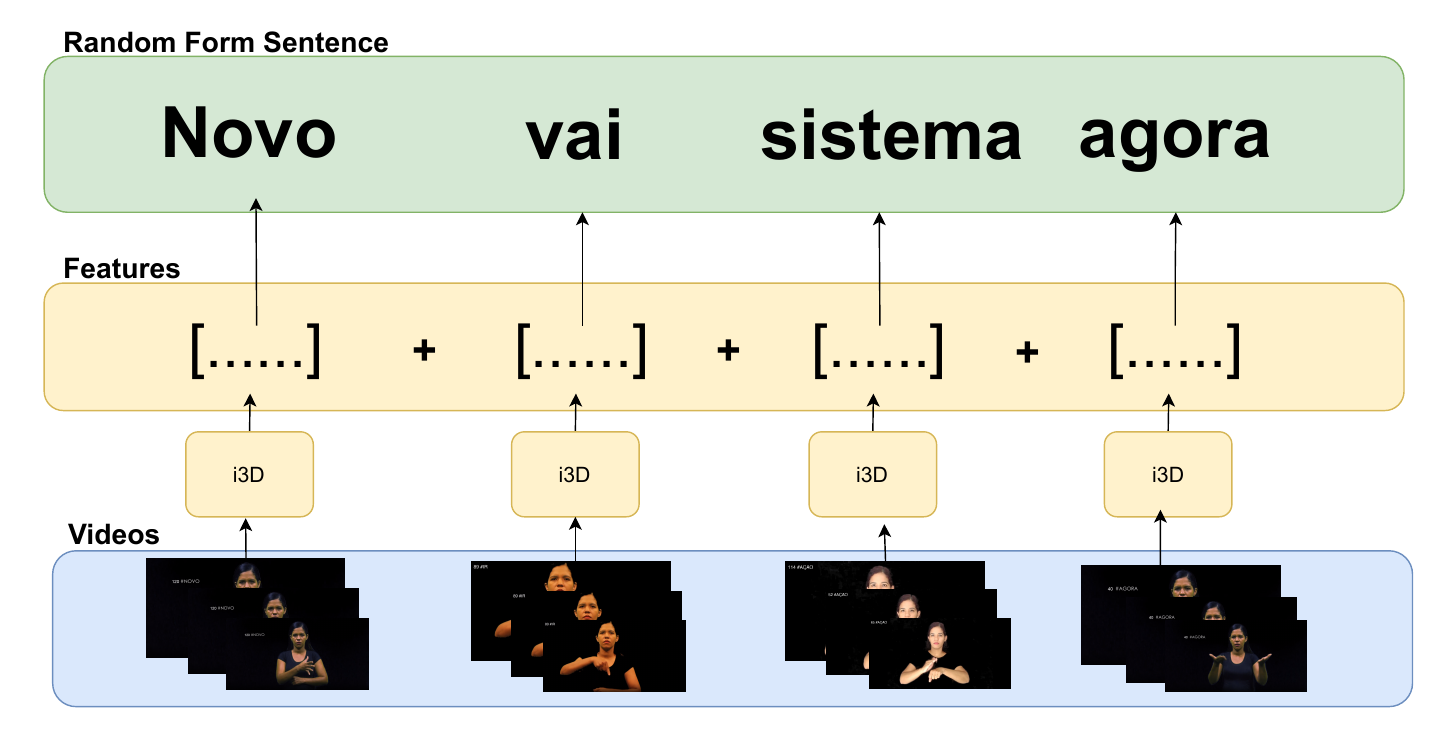} 
		}
	}    	    	    	    
    \caption{
        The process of content formation in Libras: In (a), we have an example of \gls{sf}. In (b), we have an example of sentences with \gls{rf}.
    }
   \label{fig:concatening}
\end{figure*}

We experimented with 13, 15, and 17 words per grammatical category, adding up 52, 60, and 68 words on the first, second, and third experiments, respectively.
For each phase of the experiment, the size of the training dataset was increased proportionally.
The choice to start with 52 words was based on preliminary experiments. \RV{We realized that using more than 50 words could already produce interesting performances. Our main motivation for choosing 52 was to ensure equal numbers of words per grammatical class, i.e., $13 \times 4$.
}

The sentences were crafted in both Portuguese and English. 
However, during the training phase, the English sentences were employed due \RM{to the embedding layer}\RV{to the poor performance of our prior experiments using embeddings in Portuguese}. 
The videos of the sentences were created using the same interpreter for each sentence. 
In the end, each created sentence had three different versions, corresponding to the different interpreters who signed the sentence.

V-LIBRASIL contains approximately three videos per sign, interpreted by three different people. 
During the creation of the dataset, we decided to perform the training using only two out of the three interpreters and to use the third interpreter for the validation process.
Consequently, in the training set, each sentence had 2 different versions performed by different interpreters and 4 augmented versions chosen randomly.
Thus, the same sentence \RM{would appear}\RV{appears} in the dataset 6 times.

\RV{For both experiments, the dataset size varied according to the number of words.}
\RV{For the experiment with 52 words, 171K \RM{synthetic videos}\RV{concatenated videos} of sentences were used. 
This figure is the number of all possible combinations between words from different grammatical classes, considering the 6 versions, i.e., $13^4 \times 6$.}

For the experiments with 60 words and 68 words, the proportion was used to determine the dataset sizes, resulting in approximately 300K and 500K, respectively.

The validation sets were created following these rules: for the first set, the sentences were created manually, and for the second set, they were randomly selected from the training set, but ensuring they were performed by a different interpreter.

\RV{In the experiment setup, the first validation set consisted of 52, 60, and 68 sentences for each experiment. 
The second set consisted of 100 sentences.}

It is important to highlight that the sentences in the validation set 1 were manually created to produce meaningful sentences. 
They did not follow a predefined structure and did not necessarily appear in the training set.

\subsection{STL Model}

\begin{figure*}[!ht]
\centering
\includegraphics[width=0.80\textwidth]{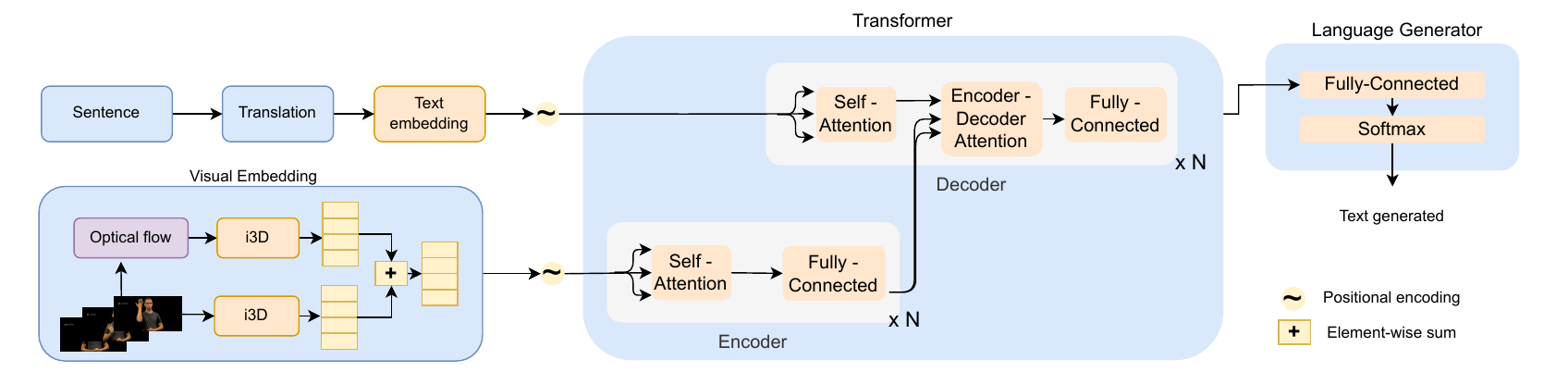}
\caption{ \RV{Overview of the Sign Language Translator Architecture used. We feed a Transformer with concatenated videos from V-LIBRASIL and with generated sentences.}}
\label{fig:architecture}
\end{figure*}

The model employed in this work utilizes the same architecture as described in~\cite{dvsilva:2023-libras}. 
Detailed information on the architecture and the underlying mathematical principles can be found in the original works~\cite{vaswani:2017attention,estevam:2021dense}. 
In our experiment, the model is fed with features from V-LIBRASIL videos, \RV{extracted using the \gls{i3d} neural network pre-trained on the Kinetics-400 dataset~\cite{carreira:2018}}, and with the sequence of tokens received from the embedding layer and derived from the generated sentences.
We use 300-dimensional GloVe vectors pre-trained on 840B tokens~\cite{pennington:2014glove} for the word embeddings. 
The features and token sequences are positionally encoded before input into the Transformer.
The language generation component, consisting of a fully connected layer followed by a softmax layer, predicts the output words \RV{as illustrated in Figure \ref{fig:architecture}.}

\RM{The experiments were conducted on a computer equipped}
\RM{with an AMD Ryzen Threadripper $1920$X $3.5$GHz CPU,}
\RM{an NVIDIA TitanXp GPU (12GB), and 96GB of RAM.}

\subsection{Evaluation Metrics}

We evaluated the translation quality using BLEU@1-4 \cite{papineni:2002bleu} and METEOR \cite{banerjee:2005meteor} metrics.
BLEU is a widely used metric for machine translation, image, and video captioning.
It compares machine translations to professional human translations using modified unigram precision.
It reports scores for n-grams (sequences of n words), with BLEU@1 focusing on single words (unigrams) and BLEU@4 considering sequences of four words.
Generally, higher BLEU scores at longer n-gram lengths indicate greater fluency.

METEOR, another popular metric, addresses limitations identified in BLEU and aims for a higher correlation with human judgment.
It uses three matching strategies: exact matches, stemmed matches (e.g., ``garden'' and ``gardens''), and synonyms from WordNet\footnote{not applicable for Portuguese evaluation}. 
We employed the script by Krishna~\cite{krishna:2017} for BLEU and METEOR calculations.

\section{Results}
\label{sec:results}

In this section, the results of the experiments involving the SF and RF approaches are described in Table \ref{table:results}, and we present a detailed analysis based on the results.
We also present qualitative results with a proper discussion.
\RV{
    The experiments were conducted on a computer equipped with an AMD Ryzen Threadripper $1920$X $3.5$GHz CPU, an NVIDIA TitanXp GPU (12GB), and 96GB of RAM.
}

\input{tables/results}

We noticed a significant performance difference between the two validation sets for the \gls{sf} configuration.
The sentence structure seems to be the major factor influencing the model's learning. \RV{We can observe this effect in Table~\ref{table:results} through the BLEU@2-4 metrics (i.e., SF-52, SF-60, and SF-68). We noticed that this significant difference is primarily due to the structural differences between the sentences in the first validation set, and the second validation set, which has different formation processes, as described in Section \ref{subsec:experimentalconf}. Based on these metrics, we observed that} \RM{We noticed that} the model could learn the content at the structural sentence level but not at the signal level, as shown by their poor performance on the first validation set (i.e., without fixed word positions). We believe the model learned the input pattern and attempted to reproduce this pattern in the output, which explains the BLEU@4 score of zero. Additionally, considering fixed positions means reducing the number of possible words in each position, making the problem easier, which explains the high performance for the validation set 2.

Following this training approach, models created with sentences based on a fixed pattern will be less consistent and have difficulty correctly translating sentences that do not follow a fixed pattern (e.g., open-world applications).
However, this type of approach could be employed to create models for translating sentences within a predictable context where the sentences that can be used are limited (e.g., Medical care, sign language teaching, basic interactions).

\input{tables/rf-sf-results}

\RV{In contrast}, the experiments under RF configuration showed that the model could learn using the random position of words to create sentences. \RV{We can observe this effect in Table \ref{table:results} through the BLEU@1 metrics (i.e., RF-52, RF-60, RF-68) due to the nature of the BLEU@1 metric, which allows us to measure the accuracy of individual words within a sentence. Based on these metrics was noted}\RM{It was  noted} that the model achieved similar scores for both validation sets. This demonstrates that the model became more consistent, learning more at the signal level rather than the sentence structure level. The greater variability in sentence formats enables the model to have better generalization capabilities.

\RV{In \gls{sf} experiment, we observed that despite the increase in the number of words from 52 to 60, the model produced a similar performance with a small difference (i.e., BLEU@1-4 in Table \ref{table:results}).}
\RV{In the RF experiment, we observed a significant performance improvement, with an increase in the number of words from 52 to 60 (i.e., BLEU@1-4 in Table \ref{table:results}), which made the problem even more difficult. Although numerically lower, it can be observed that the model performs independently of the sentence structure provided. This is an indicator that the model could be capable of learning from real data (without necessarily a fixed order of words)} 

This was achieved with the increase in the size of the training dataset. These results show that we can increase the vocabulary size while preserving or improving its translation capability. However, this increase in vocabulary and training dataset size is limited. This can be observed in the increase in vocabulary from 60 to 68 words \RV{for both experiments in Table \ref{table:results}}, where we had a decrease in all metrics scores. We argue that there is a lack of interpreter variability (i.e., only three interpreters per sign), and not displaying multiple patterns of the same sign performed by different people reduces the model's generalization capabilities. 
Indeed, our prior experiments without augmentations showed poor performance. 

Another relevant aspect is the semantics of the sentences. In the transformer architecture, words are predicted based on their context, and sentences containing words with a low probability of appearing together (i.e., our RF configuration) make learning more difficult. However, generating grammatically correct and semantically meaningful sentences from a selected set of words is not trivial and deserves attention in future works.
\RV{Moreover, the way the sentences were translated into English (i.e., merely translated word by word) is another limiting aspect of the model's performance, which can be addressed by better sentence generation strategies. }

Examples of translated sentences from our experiment 2 of \gls{rf} configuration are shown in Fig.~\ref{fig:results}. 
Fig.~\ref{fig:results}(a) and Fig.~\ref{fig:results}(b) exhibit examples of successful translation sentences, while 
Fig.~\ref{fig:results}(c) and Fig.~\ref{fig:results}(d) show examples of partially correct outputs. 
Finally, we exhibit incorrect translations yielded by the model in Fig.~\ref{fig:results}(e) and Fig.~\ref{fig:results}(f).

Analyzing Fig.~\ref{fig:results}(a), we observe that all signals were correctly identified and in the order they appear.
The importance of the augmentation procedure is highlighted by the last signal, which appears mirrored between the training and validation videos.
Another interesting aspect of Fig.~\ref{fig:results}(a) and Fig.~\ref{fig:results}(b) is related to the gender difference between the interpreters in the training and validation videos; the difference in gender does not seem to be a problem to the model. Regarding Fig.~\ref{fig:results}(c), the model can correctly recognize only the sign for ``government,'' even though it was mirrored. 
However, there is a noticeable gestural similarity between the signs for ``to go out'' and ``high'' as well as ``home'' and ``leave''.
These similarities may have confused the model, leading it to make incorrect predictions.
Regarding Fig.~\ref{fig:results}(d), the model can not recognize only the sign for ``go''. 
Although the positions and hand movements are not similar, the arm movements for the signs ``go'' and ``new'' also have similarities, confusing the model.
Regarding Fig.~\ref{fig:results}(e) and Fig.~\ref{fig:results}(f), the model failed to recognize any signs in these sentences. 
It is worth noting that in Fig.~\ref{fig:results}(f), the sign movements in the reference sentence are similar to the signs in the model prediction.
These similarities may indicate that the model cannot differentiate detailed differences for some signs.

\section{Conclusion}
\label{sec:conclusion}

In this paper, we introduced a new approach for training \gls{slt} models by \RM{synthesizing}\RV{concatenating} videos of sign sequences from short sign clips. This procedure does not require manual data labeling and enables us to generate thousands of videos.
We demonstrated the model's learning ability under different experimental configurations by changing vocabulary sizes and sentence generation strategies. 
We also show that increasing the vocabulary and dataset size allows the model to improve its performance; 
however, this improvement is limited. 
Our technique shows promising results, especially for adoption in reduced vocabulary contexts. 
In future work, we aim to explore improvements in the sentence generation mechanism and investigate more methods to produce variability in the signs, \RV{for example \cite{debem:2022}. We also intend to validate these methods with real-world videos to test the effectiveness of the generated SLT models and conduct the training without embeddings in English. }Additionally, we intend to use another channel of information, such as keypoints, for training the network.

\section*{Acknowledgment}

\iffinal
This work was partly supported by the National Council for Scientific and Technological Development (CNPq) (\# 315409/2023-1)).
We gratefully acknowledge the support of NVIDIA Corporation with the donation of the Quadro RTX 8000 GPU used for this research.
\else
We thank the anonymous.
\fi

\bibliographystyle{IEEEtran}

\bibliography{refs}

\end{document}

%% file: 0-packages.tex
\usepackage{cite}
\ifCLASSINFOpdf
   \usepackage[pdftex]{graphicx}  
   \graphicspath{{figs/}}  
   \DeclareGraphicsExtensions{.pdf,.jpeg,.png}
\else  
   \usepackage[dvips]{graphicx}  
   \graphicspath{{../figs/}}  
   \DeclareGraphicsExtensions{.eps}
\fi

\usepackage[cmex10]{amsmath}
\usepackage{amsthm}

\usepackage{algorithmic}

\usepackage{array}

\ifCLASSOPTIONcompsoc
  \usepackage[caption=false,font=normalsize,labelfont=sf,textfont=sf]{subfig}
\else
  \usepackage[caption=false,font=footnotesize]{subfig}
\fi

\usepackage{url}
\newif\iffinal
\finaltrue

\iffinal
\else
\usepackage[switch]{lineno}
\fi

\usepackage[acronym]{glossaries}

\usepackage{xcolor}
\usepackage{soul}
\usepackage{float}
\usepackage[normalem]{ulem}

%% file: 1-commands.tex
\newcommand\RV[1]{{#1}}
\newcommand\RM[1]{{}}

%% file: 2-acronyms.tex
\newacronym{libras}{Libras}{Brazilian Sign Language}
\newacronym{slt}{SLT}{Sign Language Translation}
\newacronym{SLT}{SLT}{Sign Language Translation}
\newacronym{SLR}{SLR}{Sign Language Recognition}
\newacronym{slr}{SLR}{Sign Language Recognition}
\newacronym{sf}{SF}{Structured Form}
\newacronym{rf}{RF}{Random Form}
\newacronym{SF}{SF}{Structured Form}
\newacronym{RF}{RF}{Random Form}
\newacronym{i3d}{i3D}{Inflated 3D ConvNet}

%% file: header.tex
\title{Less is more: \RM{synthesizing}\RV{concatenating} videos for Sign Language Translation from a small set of signs}

\iffinal

\author{
\IEEEauthorblockN{David Vinicius da Silva\IEEEauthorrefmark{1}, Valter Estevam\IEEEauthorrefmark{2}, and David Menotti\IEEEauthorrefmark{1}}
\IEEEauthorblockA{
  \IEEEauthorrefmark{1}Department of Informatics, Federal University of Paran\'{a}, Curitiba, Brazil \\
  \IEEEauthorrefmark{2}Federal Institute of Paraná, Irati, Brazil Brazil\\
  \resizebox{0.6\linewidth}{!}{
        \hspace{-0.75mm}\IEEEauthorrefmark{1}\hspace{-0.35mm}\tt{\small{\{david.vinicius,menotti\}}@ufpr.br} \quad \IEEEauthorrefmark{2}{\tt\small  valter.junior@ifpr.edu.br}}
 }
}

\else
  \author{SIBGRAPI Paper ID: 27 \\ }
  \linenumbers
\fi

%% file: tables/results.tex
\begin{table}[!ht]
\renewcommand{\arraystretch}{1.3}
\caption{Results of RF and SF Experiments on validation sets 1 (meaningful sentences) and 2 (randomly selected sentences).}
\label{table:results}
\centering
\resizebox{0.90\linewidth}{!}{
\begin{tabular}{|c|c|c|c|c|c|c|c|c|c|c|c|c|}
\hline
Config. & \# & \multicolumn{2}{c|}{BLEU 1} & \multicolumn{2}{c|}{BLEU 2} & \multicolumn{2}{c|}{BLEU 3} & \multicolumn{2}{c|}{BLEU 4} & \multicolumn{2}{c|}{METEOR} \\ \hline
Validation set & - & 1 & 2 & 1 & 2 & 1 & 2 & 1 & 2 & 1 & 2 \\ \hline
SF & \RM{1}\RV{52} & 26.18 & 47.99 & 5.89 & 26.68 & 1.02 & 17.79 & 0  & 9.20 & 10.59 & 26.18 \\ \hline
\RV{SF} & \RV{60} & \RV{24.50} & \RV{47.95} & \RV{5.65} & \RV{25.32} & \RV{0.51} & \RV{15.76} & \RV{0}  & \RV{8.84} & \RV{10.1} & \RV{25.22} \\ \hline
\RV{SF} & \RV{68} & \RV{20.70} & \RV{40.83} & \RV{6.32} & \RV{24.48} & \RV{1.51} & \RV{17.85} & \RV{0}  & \RV{9.39} & \RV{9.05} & \RV{21.71} \\ \hline
RF & \RM{1}\RV{52} & 39.46 & 36.06 & 15.74 & 13.69 & 4.50 & 6.19 & 2.0 & 2.01 & 18.10 & 15.21 \\ \hline
RF & \RM{2}\RV{60} & 41.88 & 37.99 & 24.38 & 19.96 & 9.23 & 8.60 & 4.18 & 4.02 & 20.72 & 17.40 \\ \hline
RF & \RM{3}\RV{68} & 28.75 & 30.70 & 7.91 & 12.45 & 2.40 & 4.68 & 0 & 1.66 & 11.96 & 14.45 \\ \hline
\end{tabular}
}
\end{table}

%% file: tables/rf-sf-results.tex
\begin{figure*}[ht]
	\centering
	\captionsetup[subfigure]{captionskip=-0.20pt,font={scriptsize},justification=centering}
	\vspace{0.45mm}
	\resizebox{0.80\linewidth}{!}{
            \subfloat[][]{
			\includegraphics[width=0.50\linewidth]{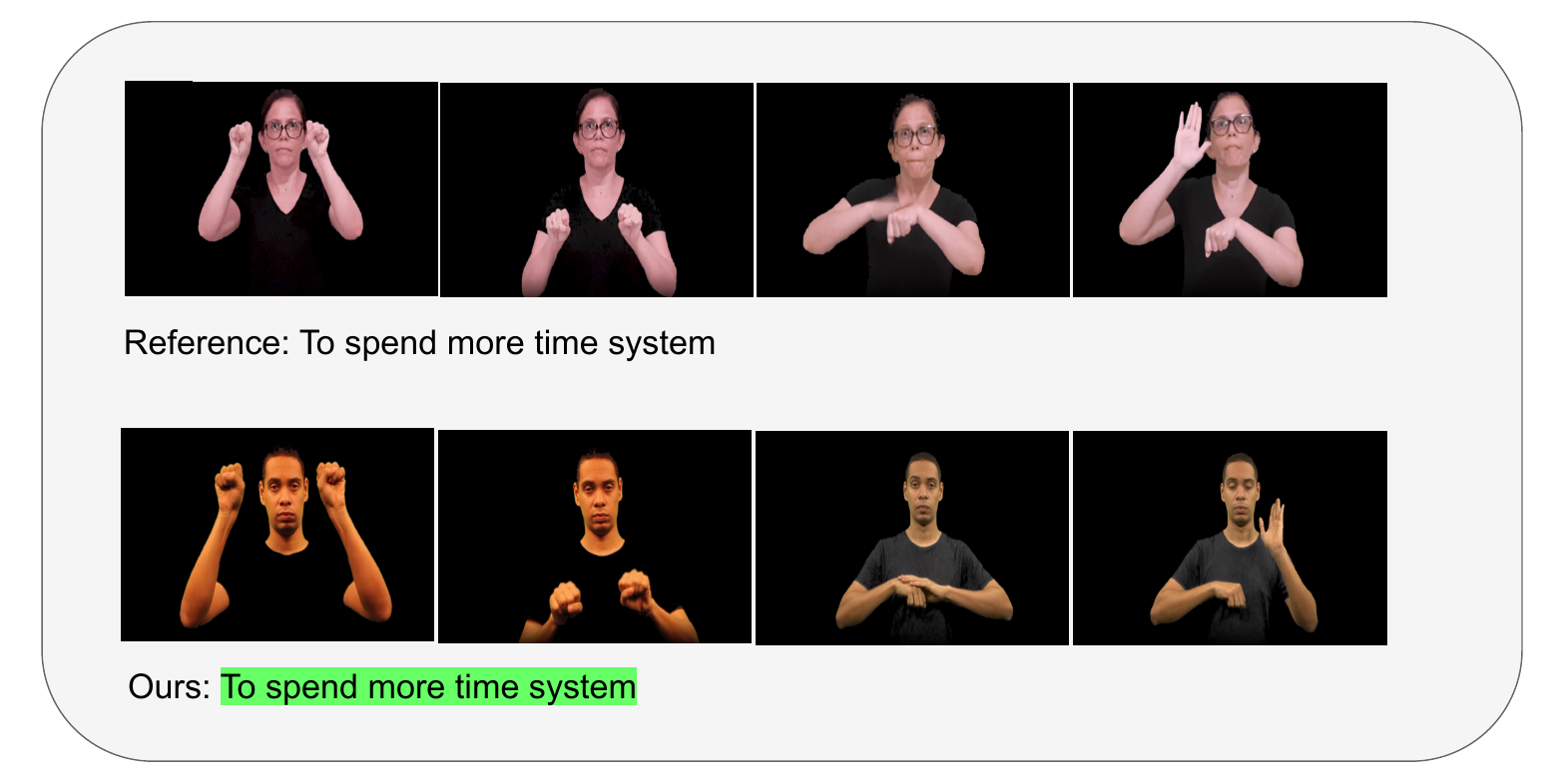} 
		}
            \subfloat[][]{
			\includegraphics[width=0.50\linewidth]{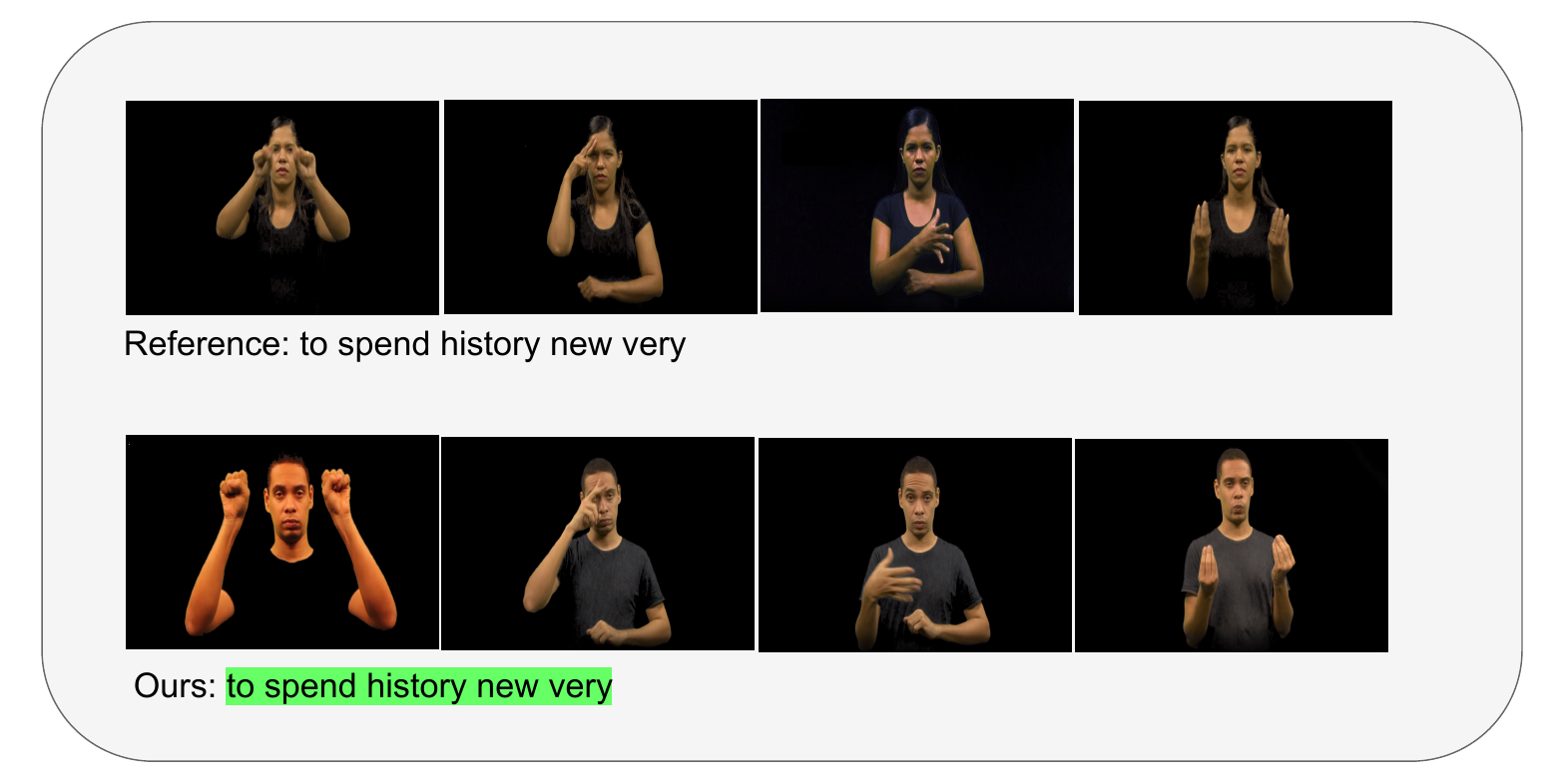} 
		}
	}    	    
	\vspace{0.45mm}
	\resizebox{0.80\linewidth}{!}{
            \subfloat[][]{
			\includegraphics[width=0.50\linewidth]{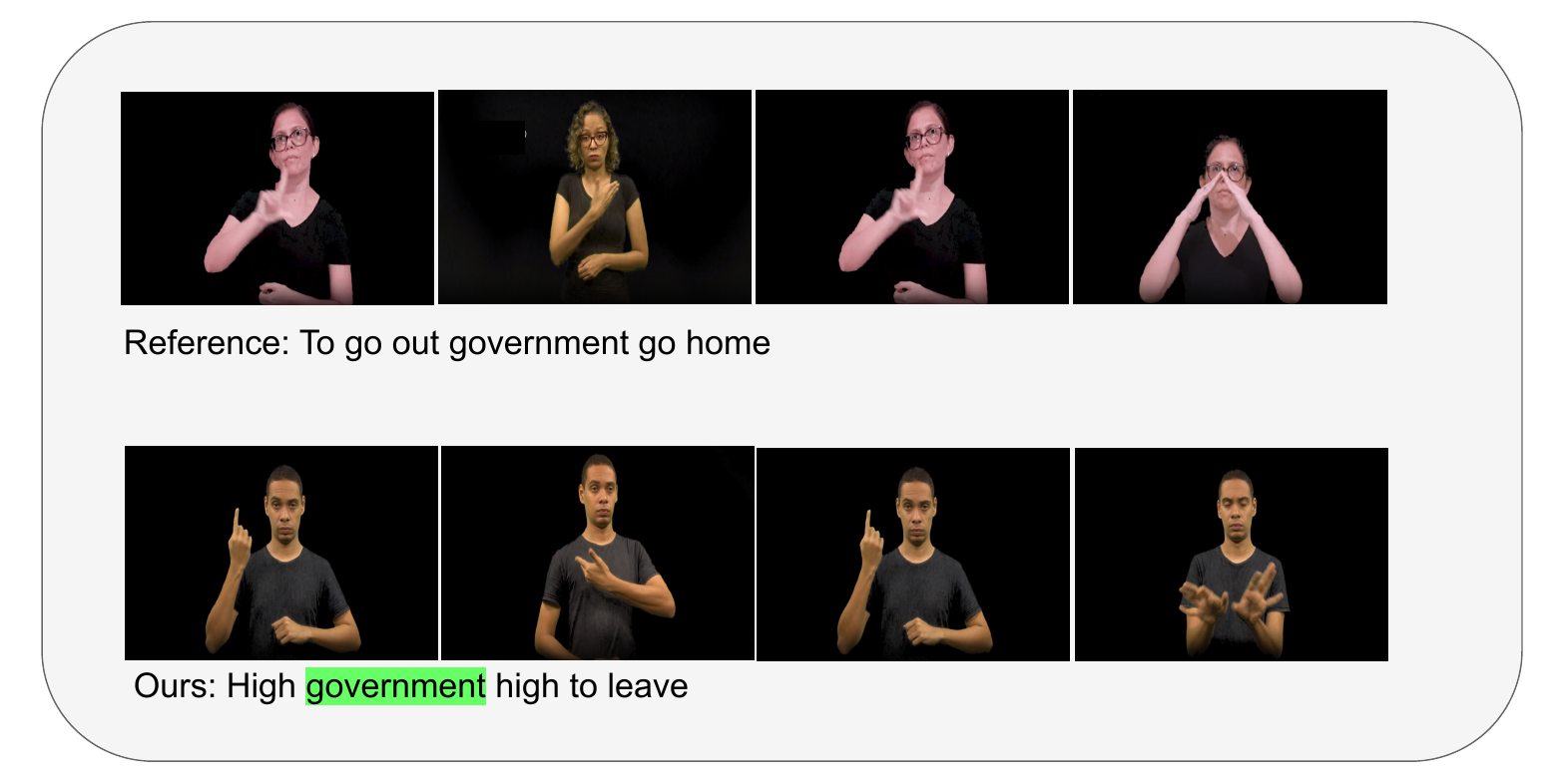} 
		}
            \subfloat[][]{
			\includegraphics[width=0.50\linewidth]{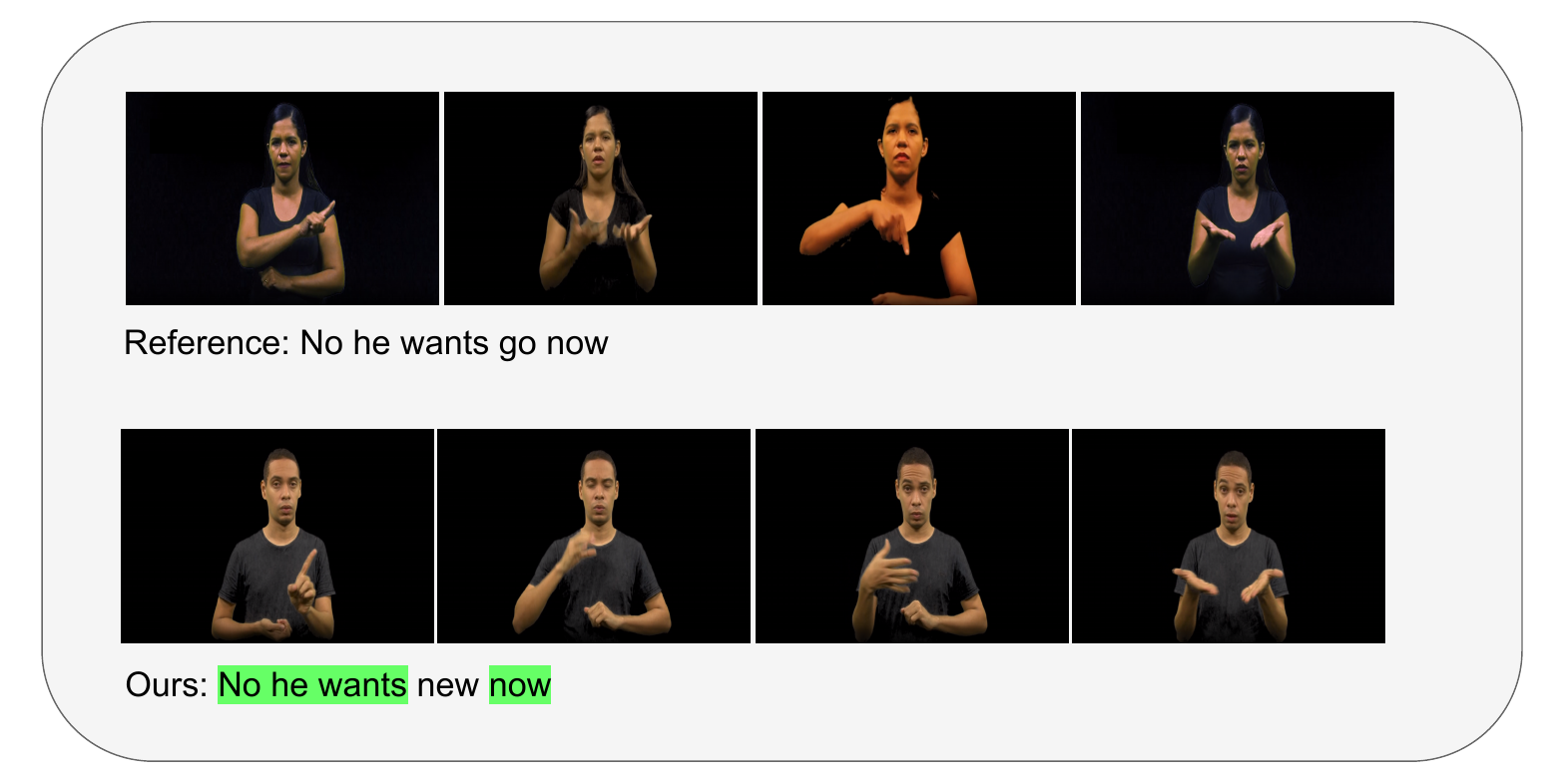} 
		}
	}    	    	   	
	\vspace{0.45mm}
    \resizebox{0.80\linewidth}{!}{
            \subfloat[][]{
			\includegraphics[width=0.50\linewidth]{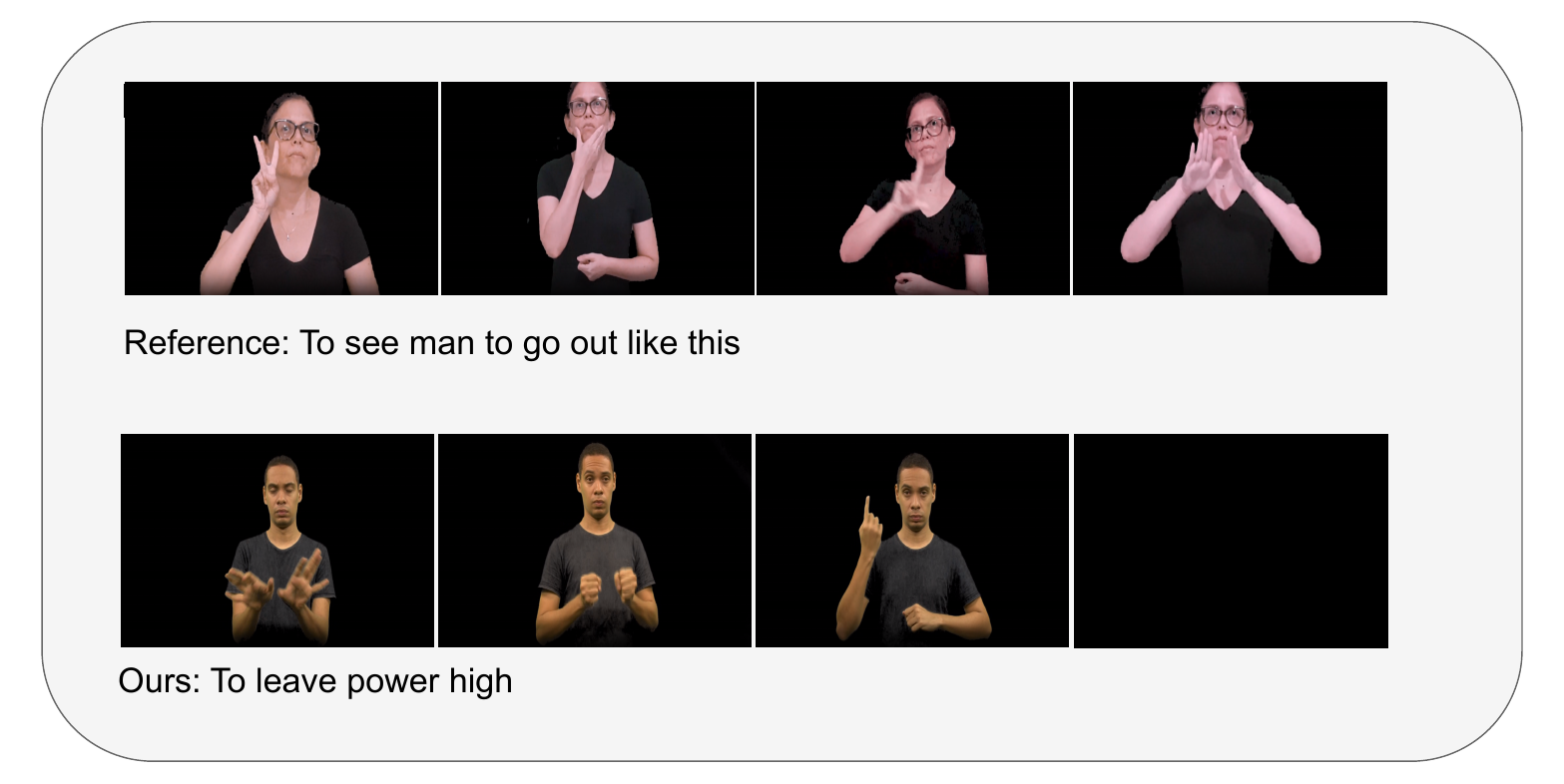} 
		}
            \subfloat[][]{
			\includegraphics[width=0.50\linewidth]{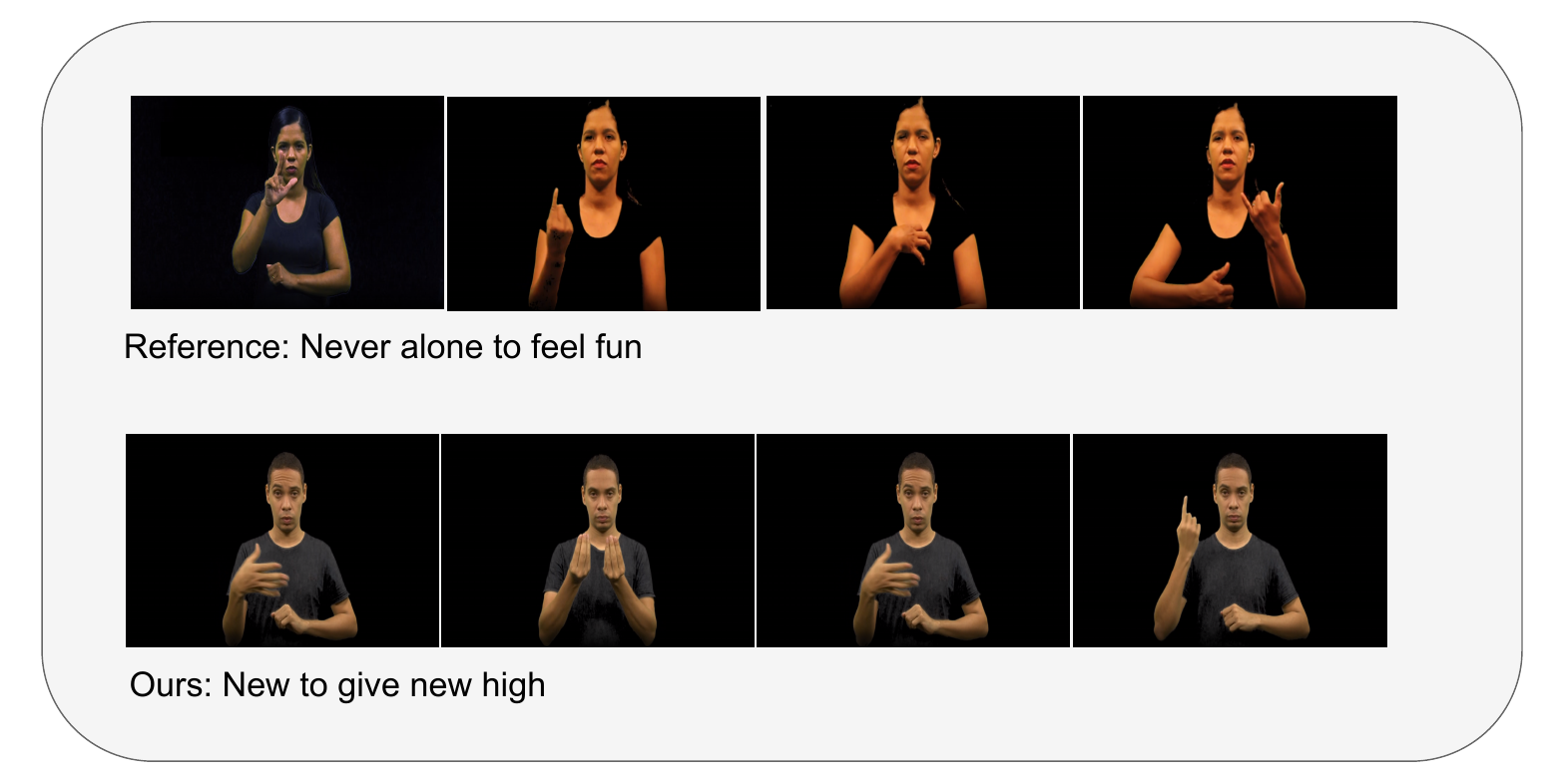} 
		}
	}    	    	   	
	\vspace{-2mm}
    \caption{
    Qualitative results: in (a) and (b), we have an example of text translated by the model using video sequences as input. The model correctly predicts all words; In (c) and (d), we have examples of parcials corrects outputs generated by the model; in (e) and (f), we can see examples where the model fail to translate the sentences.
    }
   \label{fig:results}
\end{figure*}

%% file: main.bbl
\begin{thebibliography}{10}
\providecommand{\url}[1]{#1}
\csname url@samestyle\endcsname
\providecommand{\newblock}{\relax}
\providecommand{\bibinfo}[2]{#2}
\providecommand{\BIBentrySTDinterwordspacing}{\spaceskip=0pt\relax}
\providecommand{\BIBentryALTinterwordstretchfactor}{4}
\providecommand{\BIBentryALTinterwordspacing}{\spaceskip=\fontdimen2\font plus
\BIBentryALTinterwordstretchfactor\fontdimen3\font minus
  \fontdimen4\font\relax}
\providecommand{\BIBforeignlanguage}[2]{{%
\expandafter\ifx\csname l@#1\endcsname\relax
\typeout{** WARNING: IEEEtran.bst: No hyphenation pattern has been}%
\typeout{** loaded for the language `#1'. Using the pattern for}%
\typeout{** the default language instead.}%
\else
\language=\csname l@#1\endcsname
\fi
#2}}
\providecommand{\BIBdecl}{\relax}
\BIBdecl

\bibitem{leiLibras}
\BIBentryALTinterwordspacing
Brasil, ``Lei nº 10.436, de 24 de abril de 2002.'' \emph{Diário Oficial [da]
  República Federativa do Brasil}, 2002. [Online]. Available:
  \url{http://www.planalto.gov.br/ccivil_03/Leis/2002/L10436.htm}
\BIBentrySTDinterwordspacing

\bibitem{decretoLibras}
\BIBentryALTinterwordspacing
------, ``Decreto nº 5.626, de 22 de dezembro de 2005,'' \emph{Diário Oficial
  [da] República Federativa do Brasil}, 2005. [Online]. Available:
  \url{http://www.planalto.gov.br/ccivil_03/_ato2004-2006/2005/decreto/d5626.htm}
\BIBentrySTDinterwordspacing

\bibitem{rezende:2021reconhecimento}
T.~M. Rezende, ``Reconhecimento automático de sinais da libras:
  desenvolvimento da base de dados {MINDS-Libras} e modelos de redes
  convolucionais,'' Ph.D. dissertation, Universidade Federal de Minas Gerais,
  2021.

\bibitem{gameiro:2020signLanguageDataset}
P.~V. Gameiro, W.~L. Passos, G.~M. Araujo, A.~A. de~Lima, J.~N. Gois, and A.~R.
  Corbo, ``A brazilian sign language video database for automatic
  recognition,'' in \emph{2020 Latin American Robotics Symposium (LARS), 2020
  Brazilian Symposium on Robotics (SBR) and 2020 Workshop on Robotics in
  Education (WRE)}, 2020, pp. 1--6.

\bibitem{passos:2021gait}
W.~L. Passos, G.~M. Araujo, J.~N. Gois, and A.~A. de~Lima, ``A gait energy
  image-based system for brazilian sign language recognition,'' \emph{IEEE
  Transactions on Circuits and Systems I: Regular Papers}, vol.~68, no.~11, pp.
  4761--4771, 2021.

\bibitem{Camgoz:2018:NSLT}
N.~C. Camgoz, S.~Hadfield, O.~Koller, H.~Ney, and R.~Bowden, ``Neural sign
  language translation,'' in \emph{2018 IEEE/CVF Conference on Computer Vision
  and Pattern Recognition}, 2018, pp. 7784--7793.

\bibitem{dvsilva:2023-libras}
D.~V. da~Silva, V.~Estevam, and D.~Menotti, ``Towards a realistic libras to
  portuguese translation,'' in \emph{2023 36th SIBGRAPI Conference on Graphics,
  Patterns and Images (SIBGRAPI)}, 2023, pp. 1--6.

\bibitem{zhou:2021improving}
H.~Zhou, W.~Zhou, W.~Qi, J.~Pu, and H.~Li, ``Improving sign language
  translation with monolingual data by sign back-translation,'' 2021.

\bibitem{rodrigues:2021v}
A.~J. Rodrigues, ``{V-LIBRASIL}: uma base de dados com sinais na l{\'\i}ngua
  brasileira de sinais ({Libras}),'' Master's thesis, Universidade Federal de
  Pernambuco, 2021.

\bibitem{chen:2023twostream}
Y.~Chen, R.~Zuo, F.~Wei, Y.~Wu, S.~Liu, and B.~Mak, ``Two-stream network for
  sign language recognition and translation,'' 2023.

\bibitem{guan:2024multistream}
M.~Guan, Y.~Wang, G.~Ma, J.~Liu, and M.~Sun, ``Multi-stream keypoint attention
  network for sign language recognition and translation,'' 2024.

\bibitem{nltk:2002}
\BIBentryALTinterwordspacing
E.~Loper and S.~Bird, ``Nltk: The natural language toolkit,'' 2002. [Online].
  Available: \url{https://arxiv.org/abs/cs/0205028}
\BIBentrySTDinterwordspacing

\bibitem{vidaug:2018}
O.~Kopuklu, N.~Kose, A.~Gunduz, and G.~Rigoll, ``vidaug,''
  \url{https://github.com/okankop/vidaug}, 2018.

\bibitem{carreira:2018}
J.~Carreira and A.~Zisserman, ``Quo vadis, action recognition? a new model and
  the kinetics dataset,'' in \emph{2017 IEEE Conference on Computer Vision and
  Pattern Recognition (CVPR)}.\hskip 1em plus 0.5em minus 0.4em\relax IEEE
  Computer Society, 2017, pp. 4724--4733.

\bibitem{xing:2023svformer}
Z.~Xing, Q.~Dai, H.~Hu, J.~Chen, Z.~Wu, and Y.-G. Jiang, ``Svformer:
  Semi-supervised video transformer for action recognition,'' in
  \emph{Proceedings of the IEEE/CVF Conference on Computer Vision and Pattern
  Recognition (CVPR)}, June 2023, pp. 18\,816--18\,826.

\bibitem{estevam:2021dense}
V.~Estevam, R.~Laroca, H.~Pedrini, and D.~Menotti, ``Dense video captioning
  using unsupervised semantic information,'' \emph{arXiv - 2112.08455}, 2021.

\bibitem{sun:2018pwc}
D.~Sun, X.~Yang, M.-Y. Liu, and J.~Kautz, ``{PWC-Net: CNNs} for optical flow
  using pyramid, warping, and cost volume,'' in \emph{IEEE Conference on
  Computer Vision and Pattern Recognition}, 2018, pp. 8934--8943.

\bibitem{vaswani:2017attention}
A.~Vaswani, N.~Shazeer, N.~Parmar, J.~Uszkoreit, L.~Jones, A.~N. Gomez,
  L.~Kaiser, and I.~Polosukhin, ``Attention is all you need,'' in \emph{31st
  International Conference on Neural Information Processing Systems}, ser.
  NIPS'17.\hskip 1em plus 0.5em minus 0.4em\relax Red Hook, NY, USA: Curran
  Associates Inc., 2017, p. 6000–6010.

\bibitem{pennington:2014glove}
\BIBentryALTinterwordspacing
J.~Pennington, R.~Socher, and C.~Manning, ``{G}lo{V}e: Global vectors for word
  representation,'' in \emph{Proceedings of the 2014 Conference on Empirical
  Methods in Natural Language Processing ({EMNLP})}.\hskip 1em plus 0.5em minus
  0.4em\relax Doha, Qatar: Association for Computational Linguistics, Oct.
  2014, pp. 1532--1543. [Online]. Available:
  \url{https://aclanthology.org/D14-1162}
\BIBentrySTDinterwordspacing

\bibitem{papineni:2002bleu}
\BIBentryALTinterwordspacing
K.~Papineni, S.~Roukos, T.~Ward, and W.-J. Zhu, ``{B}leu: a method for
  automatic evaluation of machine translation,'' in \emph{40th Annual Meeting
  of the Association for Computational Linguistics}.\hskip 1em plus 0.5em minus
  0.4em\relax Philadelphia, Pennsylvania, USA: Association for Computational
  Linguistics, Jul. 2002, pp. 311--318. [Online]. Available:
  \url{https://aclanthology.org/P02-1040}
\BIBentrySTDinterwordspacing

\bibitem{banerjee:2005meteor}
\BIBentryALTinterwordspacing
S.~Banerjee and A.~Lavie, ``{METEOR}: An automatic metric for {MT} evaluation
  with improved correlation with human judgments,'' in \emph{{ACL} Workshop on
  Intrinsic and Extrinsic Evaluation Measures for Machine Translation and/or
  Summarization}.\hskip 1em plus 0.5em minus 0.4em\relax Ann Arbor, Michigan:
  Association for Computational Linguistics, Jun. 2005, pp. 65--72. [Online].
  Available: \url{https://aclanthology.org/W05-0909}
\BIBentrySTDinterwordspacing

\bibitem{krishna:2017}
R.~Krishna, K.~Hata, F.~Ren, L.~{Fei-Fei}, and J.~C. Niebles,
  ``Dense-captioning events in videos,'' in \emph{International Conference on
  Computer Vision (ICCV)}, 2017, pp. 706--715.

\bibitem{debem:2022}
W.~Silveira, A.~Alaniz, M.~Hurtado, B.~C. Da~Silva, and R.~De~Bem, ``Synlibras:
  A disentangled deep generative model for brazilian sign language synthesis,''
  in \emph{2022 35th SIBGRAPI Conference on Graphics, Patterns and Images
  (SIBGRAPI)}, vol.~1, 2022, pp. 210--215.

\end{thebibliography}
